\documentclass[journal,cmex10]{IEEEtran}

\usepackage{url}

\usepackage[margin=1in]{geometry}  
\usepackage{graphicx}          
\usepackage{pdfpages}

\usepackage{breqn}
\usepackage{amsfonts, amssymb, amsthm}
\usepackage{enumerate}  
\usepackage{graphics, epsfig, color}
\usepackage{xcolor}
\usepackage[caption=false,font=footnotesize]{subfig}
\usepackage{wrapfig}
\usepackage{cite}

\usepackage{authblk}
\usepackage{multirow}
\usepackage{tikz}
\usepackage{comment}
\usepackage[T1]{fontenc}  
\usepackage{lmodern}      
\usepackage{amsmath}

\usetikzlibrary{positioning}

\usepackage{calrsfs}

\def\p{\partial}
\def\non{\nonumber}

\def\nab{\nabla}

\def\tr{\rm tr}

\def\ep{\epsilon}

\newtheorem{theorem}{Theorem}[section]

\newtheorem{prop}[theorem]{Proposition}

\newtheorem{remark}[theorem]{Remark}
\theoremstyle{definition}

\graphicspath{{figures/}}

\title{\textsc{Optimal Initialization Scale for Neural Networks with Locally Quadratic Loss Landscapes: An SGD Dynamics Perspective}} 
\author[1,2]{Hiroshi Horii}
\author[3]{Sothea Has}
\affil[1]{Laboratoire Jacques-Louis Lions, Sorbonne Université, Université de Paris et CNRS, Paris, France}
\affil[2]{Université Côte d’Azur, CNRS, Observatoire de la Côte d’Azur, Laboratoire Lagrange, Nice, France}
\affil[3]{Research and Innovation Center, Institute of Technology of Cambodia, Russian Federation Blvd., P.O. Box 86, Phnom Penh, Cambodia}

\begin{document}

\makeatletter
\def\@IEEEpubidpullup{5.5\baselineskip}
\makeatother

\IEEEpubid{%
\parbox{\textwidth}{\scriptsize
\copyright~2026 IEEE. Personal use of this material is permitted.
Permission from IEEE must be obtained for all other uses, in any current
or future media, including reprinting/republishing this material for
advertising or promotional purposes, creating new collective works,
for resale or redistribution to servers or lists, or reuse of any
copyrighted component of this work in other works.\\[2pt]
H. Horii and S. Has,
``Optimal Initialization Scale for Neural Networks with Locally Quadratic
Loss Landscapes: An SGD Dynamics Perspective,''
\textit{IEEE Transactions on Neural Networks and Learning Systems},
doi: 10.1109/TNNLS.2026.3721153.\\[2pt]
Available: \url{https://ieeexplore.ieee.org/document/11657466}
}}

\maketitle

\begin{abstract}
Stochastic gradient descent (SGD), one of the most fundamental optimization algorithms in machine learning (ML), can be recast through a continuous-time approximation as a Fokker-Planck equation for Langevin dynamics, a viewpoint that has motivated many theoretical studies. Within this framework, we study the relationship between the quasi-stationary distribution derived from this equation and the initial distribution through the Kullback-Leibler (KL) divergence. As the quasi-steady-state distribution depends on the expected cost function, the KL divergence eventually reveals the connection between the expected cost function and the initialization distribution. Under the assumptions of a locally quadratic neural-network loss landscape and a quasi-stationary SGD regime, we derive an explicit upper bound for the expected loss in terms of the initialization variance. Then, by minimizing this bound, we obtain an optimal condition of the initialization variance in the Gaussian case. This result provides a concrete mathematical criterion, rather than a heuristic approach, to select the scale of weight initialization within the locally quadratic SGD regime. In addition, we experimentally confirm our theoretical results by using classical SGD to train shallow fully connected neural networks on the MNIST dataset and a two-layer CNN model on the CIFAR-10 dataset. The results show that, for the simple network model, if the variance of the initialization distribution satisfies our theoretical optimal condition, then the corresponding network achieves lower final training loss and higher test accuracy than the conventional He-normal initialization. For the two-layer CNN model, He-normal initialization gives the best performance among the tested initializations. This observation is consistent with the proposed optimal condition, since the variance ratio of He-normal initialization remains close to the optimal value not only at the overall level but also approximately at the layer-wise level. This suggests that satisfying the proposed condition at both global and layer-wise scales may be important for achieving lower loss in this architecture. We also discuss the limitation of our optimal condition on complex networks. Our work thus supplies a mathematically grounded indicator for choosing the initialization variance and clarifies its physical meaning in terms of the parameter dynamics in neural network models.

\end{abstract}
\IEEEpubidadjcol

\section{Introduction}
Stochastic Gradient Descent (SGD) is a cornerstone optimization algorithm that underpins many advancements in machine learning (ML) and deep learning (DL) \cite{wang2019deep,zhan2024simultaneous ,song2025black }. In general, if $L$ is a differentiable loss function of any ML model with trainable parameters $W\in\mathbb{R}^K$ (all weights and biases of neural networks, for example), then training such a model is equivalent to searching for the optimal parameter $W^*\in\mathbb{R}^K$ that minimizes $L$, that is,

\begin{equation}
    L(W^*) :=\inf_{W \in \mathbb{R}^K}L(W).
\end{equation}
SGD approximates such $W^*$ by generating a sequence of parameter $(W^{(t)})_{t\geq 0}$ such that $W^{(t)}\approx W^*$ for sufficiently large $t$. The main recurrence relation in SGD is given by
\begin{equation}
\label{eq:sgd_step}
    W^{(t+1)}:=W^{(t)}-\alpha \nabla \hat{L}(W^{(t)}),
\end{equation}
where $\alpha$ is the \textit{learning rate}, $\hat{L}$ is the estimated loss function computed using a small subset of training data called \textit{minibatch}, and $\nabla \hat{L}$ is the gradient of $\hat{L}$ with respect to the weight $W$. In practice, the initialization $W^{(0)}\sim p_0$ for some probability distribution $p_0$ (the common choice is Gaussian distribution $\mathcal{N}(0,\sigma_0^2 I)$ for some $\sigma_0>0$). At each step, the algorithm tries to move toward the direction with the steepest loss function at a constant rate of $\alpha>0$.

Despite its widespread applications, the theoretical understanding of its dynamics and behavior remains an active area of research. For instance, several theoretical studies have been introduced to explore the convergence property of SGD \cite{rakhlin2011making}, the convergence around flat minima \cite{hochreiter1997flat,keskar2016large,dziugaite2017computing}, and heavy-tailed phenomena in SGD \cite{ gurbuzbalaban2021heavy,hodgkinson2021multiplicative, simsekli2020hausdorff, lehman2024approximate}. Other studies focus on the influence of hyperparameters, including the learning rate $\alpha$ and the batch size $b$, on the behavior of SGD \cite{smith2017don, keskar2016large}. In addition, the choice of weight‐initialization strategy also has a significant impact on both convergence behavior and generalization performance of the model as well (see, for example, \cite{mishkin2015all,arpit2019initialize }). In particular, the widely used Xavier initialization \cite{glorot2010understanding} and He‐normal initialization \cite{he2015delving} have proven effective in practice for efficient learning in deep neural networks. Additionally, \cite{hanin2018start} briefly discusses how initialization choices are related to network architecture.

This study focuses on analyzing the impact of the initialization parameter using the continuous-time approximation method of SGD. First, we approximate the stochastic sequence generated by equation \eqref{eq:sgd_step} using a continuous-time framework. This approximation is well established to study the relation between minibatch size and the learning rate in Artificial Neural Networks (ANNs), for example, by \cite{smith2017don, goyal2017accurate, hoffer2017train,li2019stochastic}. Next, by applying the Fokker-Planck equation, we derived the steady-state distribution of the converged process under standard conditions of the loss function $L$. Several recent studies have cast SGD dynamics in terms of a Fokker–Planck formalism (e.g.\ \cite{dai2020large,mandt2017stochastic,ziyin2025noise,chaudhari2018stochastic}). In particular,\cite{ziyin2025noise} suggested that the stationary state of SGD can retain dependence on the initialization through a memory effect. Moreover, \cite{mandt2017stochastic} connected the analytically derived steady state of the associated diffusion process with the empirically observed quasi–steady state by explicitly quantifying their KL divergence. These perspectives are closely related to the methodology developed in our study. However, based on the continuous-time approximation, we establish the main additional contributions, as listed below.
\begin{itemize}
  \item We derive, through a Kullback-Leibler divergence analysis, the analytical bound for the expected loss function with Gaussian weight initialization in the classical SGD.
  \item We show that for small initialization variances (\(\sigma_0^2 \ll 1\)), the expected loss scales as \(\mathbb{E}[L(W)] = O\bigl(\mathbb{V}(W)\bigr)\), where $\mathbb{V}(W)$ is the variance of the parameter $W$ at quasi-steady state. Meanwhile, for large variances (\(\sigma_0^2 \gg 1\)), it behaves like \(O(\log \sigma_0^2)\).
  \item By minimizing the upper bound derived from the expected loss function with respect to \(\sigma_0^2\), we obtain an optimal condition for the initialization variance $\sigma_0^2$.
  \item We validate our theoretical result through numerical experiments by comparing it against the widely used He-normal initialization. It shows the existence of an optimal hyperparameter in three-layer deep neural network models (DNNs), and a new perspective that confirms our optimal condition can also be observed in a two-layer CNN model initialized using the He-normal distribution.
  \item Finally, our numerical experiments seem to suggest that better performance is obtained when the optimal condition is satisfied not only for overall layer weights but also for the parameters at each individual layer.
  
\end{itemize}

This paper is organized as follows. Section~\ref{sec:SDE_of_SGD} provides the background on our work, focusing on the continuous-time approximation of the SGD sequence and deriving its steady-state distribution. The main result of this study is presented in Section~\ref{sec:reg_of_SGD}. Numerical experiments carried out on the MNIST and CIFAR-10 data sets are provided in Section~\ref{sec:numerical}. Finally, Section~\ref{sec:conclusion} concludes the result of this study and discusses possible directions for future investigations.

\section{Stochastic differential equation framework for SGD}
\label{sec:SDE_of_SGD}
\subsection{Continuous-time approximation for SGD}
In this section, we provide the main theoretical framework for understanding the optimization problems of the SGD in ML. The SGD is one of the most widely adopted and basic optimization algorithms. In addition, the SGD is denoted as a discretized recurrence equation for an optimized parameter in ML. Its time evolution is driven by the loss function gradient, and the step size is defined by the learning rate $\alpha>0$. In the following, let $N$ denote the total size of the training data. Under the condition that the ratio $\alpha/N$ is sufficiently small, we can obtain a stochastic differential equation (SDE) by rewriting \eqref{eq:sgd_step} as follows.

\begin{align}
	W^{(t+1)}-W^{(t)} &= -\frac{\alpha}{N} \left[\nab L +(N\nab  \hat{L}-\nab L )\right]\nonumber\\
 &=-\frac{\alpha}{N} \left[\nab L +(\nab  \tilde{L}-\nab L )\right],
\end{align}	
 where $\nab L=\sum^N_i\nab L_i$ is the sum of the true gradient $\nab L_i$ evaluated at observation $i$, and $\nab  \tilde{L}=\frac{N}{b}\sum_{i\in\Omega_b} \nab L_i$ is an estimated gradient evaluated by multiplying the average individual gradients (over a minibatch) with the sample size $N$. Intuitively, $\nab \tilde{L}$ approaches $\nab L$ as $b$ approaches $N$. Then, the equation is rewritten as follows.
\begin{equation}\label{eq:Langevin}
	W^{(t+1)}-W^{(t)} = -\frac{\alpha}{N} \left[\nab L -\eta(t) \right].
\end{equation}	
Here, $\eta(t) = \nab L-\nab  \tilde{L}$ is the difference between the true gradient and the sampled gradient at $W^{(t)}$. It reflects the error introduced when approximating the true gradient of the loss function using a minibatch gradient. If the batch size is sufficiently large, the central limit theorem can be assumed to hold for this gradient error. Consequently, the expected value of $\eta$ is $\mathbb{E}[\eta]=0$, and the gradient error can be seen as Gaussian noise for the recurrence equation. Under this condition, the noise is characterized by $\mathbb{E}[\eta(t)\eta(t')]=N(\frac{N}{b}-1) \Sigma\delta(t-t')\approx \frac{N^2}{b}\Sigma\delta(t-t')$, where $\Sigma$ is a covariance matrix \cite{smith2017don} when $N$ is sufficiently large compared to $b$. We assume that this covariance matrix has a square root decomposition formula, which is given by $\Sigma=BB^T$. Furthermore, if $\alpha$ is sufficiently small and comparable to the size of the discretization time step $\Delta t$ $(\Delta t:=\alpha)$, then the continuous-time approximation can be applied. In the regime where $\alpha$ is sufficiently small, this equation can be interpreted as a stochastic differential equation (SDE), expressed as follows:
\begin{equation}
	dW^{(t)} = -\nab L dt + \sqrt{\ep }B d V_t.
	\label{eq:SDE}
\end{equation}	
where $\ep$ is $\frac{\alpha}{b}$. In addition, $V(t)$ follows $V(t)\sim\mathcal{N}(0,1)$. Note that the coefficient of the stochastic term $\ep^{\frac{1}{2}}$ is obtained by considering $\lim_{\Delta t\rightarrow 0}\sqrt{\Delta t} \eta = dV_t$. This equation can also be viewed as an equation for the Wiener process.

\subsection{Fokker-Planck equation}\label{sec:fokker-planck}
Under the small regime of $\ep$, the SGD can be seen as an SDE with the Wiener process. Thus, by using Ito calculus, we can derive the Fokker-Planck equation for the SGD framework. The Fokker-Planck equation allows us to describe the evolution of the probability distribution function over time, which is useful for understanding the distributional dynamics of weights in SGD. 

Suppose we have a test function $Y^{(t)}=\phi(W^{(t)})=(Y_k)$, Ito calculus for the $k$th stochastic process component $Y_k$, is calculated as
\begin{equation}
	dY_k^{(t)}=\sum_i \frac{\p \phi_k}{\p w_i}dW_i^{(t)} + \frac{1}{2}\sum_i\sum_j\frac{\p^2 \phi_k}{\p w_iw_j}dW_i^{(t)}dW_j^{(t)}.
\end{equation}
Finally, we get
\begin{align}
	d \phi_k= \sum_i \frac{\p \phi_k}{\p w_i}\left(-\nab L dt + \sqrt{\ep  }Bd V_t\right)_i \non\\+ \frac{1}{2}\sum_i\sum_j\frac{\p^2 \phi_k}{\p w^2_i}(\ep\Sigma)\delta_{ij}dt\non\\
	\frac{d \phi_k}{dt}= \sum_i \frac{\p \phi_k}{\p w_i}\left(-\nab L + \sqrt{\ep  }B\frac{d V_t}{dt}\right)_i\non\\+ \frac{1}{2}\sum_i\sum_j\frac{\p^2 \phi_k}{\p w^2_i}(\ep\Sigma)\delta_{ij}.\label{eq:ito}
\end{align}
Here, we use $dV_idV_j=\delta_{ij}dt$, where $\delta_{ij}$ is Kronecker delta. To obtain the Fokker-Planck equation (Kolmogorov forward equation), we take the expectation of \eqref{eq:ito}:
\begin{align}
\mathbb{E}\!\biggl[\frac{d\phi_k}{dt}\biggr]
&=\;\mathbb{E}\!\biggl[\,
    \sum_i \frac{\partial \phi_k}{\partial w_i}
      \Bigl(-\nabla L + \sqrt{\varepsilon}\,B\,\tfrac{dV_t}{dt}\Bigr)_i
    \notag\\[-0.5ex]
&\quad\;+\;\tfrac12\sum_{i,j}
      \frac{\partial^2 \phi_k}{\partial w_i\,\partial w_j}
      \,\varepsilon\,\Sigma\,\delta_{ij}
  \,\biggr].
\end{align}
Then, we obtain
\begin{align}
\frac{\partial p(W,t)}{\partial t}
&= \sum_i \frac{\partial}{\partial w_i}
      \bigl(p(W,t)\,\nabla L \bigr)_i
   \notag\\[-0.5ex]
&\quad
  + \tfrac12 \sum_{i,j}
      \frac{\partial^2}{\partial w_i\,\partial w_j}
      \bigl(p(W,t)\,\varepsilon\,\Sigma \bigr)\,\delta_{ij}\,.\label{eq:first_fokker_eq}
\end{align}
Equation \eqref{eq:first_fokker_eq} encodes the dynamics of the weight distribution function and is useful for understanding its behavior.
By using a different stochastic process, such as "Levy-process", we can also derive a similar SDE \cite{simsekli2019tail}. Note that we used Gaussian noise to derive the Fokker-Planck equation, and this noise comes from the gradient error between the true gradient and the gradient on the minibatch learning. 

Our theory is developed around the quasi-convergence of SGD. Therefore, a usual assumption for the existence of such a state is required and stated in the following part. We first assume that around the deep local minimum, the loss function is well approximated by a quadratic form, which is given by
\begin{equation}
\label{eq:quasi_loss}
    L(W)=\frac{1}{2} W^{T}AW,
\end{equation}
where $A$ is positive definite and $\nab L=0$. Without loss of generality, we assume that a minimum loss is at $W$ = 0. Through this assumption, we can describe Gibbs-like quasi-steady state distribution, and it makes sense when the loss function is smooth and the stochastic process reaches a low-variance quasi-stationary distribution around a deep local minimum\cite{mandt2017stochastic}. Moreover, within the small regime of learning rate, the exit time from the local minimum is very long \cite{kramers1940brownian}. Thus, the local minimum is deep, and a quasi-steady state exists. It is given by solving the Fokker-Planck equation when we assume the distribution converges to a steady state after exploring the state space long enough, under isotropic variance, i.e., $\Sigma=\sigma^2 I$. At the quasi-steady state, the solution $p(W)$ evolves independently of time $t$. By applying this condition on \eqref{eq:first_fokker_eq}, we obtain the following equation:
\begin{equation}
	\nab p(W) = -p(W)\frac{2}{\ep \sigma^2}\nab L.
\end{equation}
Therefore, the solution of the weight distribution function is given by
\begin{equation}\label{eq:ss}
	p(W)=C_{1,\text{loc}}e^{-\frac{2bL(W)}{\alpha\sigma^2}}.
\end{equation}
Here, we use $\frac{\alpha}{b}=\ep$, and $C_{1,\text{loc}}$ is a normalized constant for a distribution for one local minimum. This formula contains the structure of the loss function and output softmax function through the loss function $L$. By collecting all the necessary conditions, the following assumptions are made throughout this study:
\begin{enumerate}
    \item The learning rate $\alpha$ is sufficiently small, and SGD can be written as an SDE in continuous time approximation.
    \item The covariance matrix is an isotropic structure $\Sigma=\sigma^2 I$.
    \item This stochastic dynamics goes to a deep local minimum with the quadratic loss function, and the escape time is sufficiently large. Then, we have a quasi-steady state for the Fokker-Planck formalization.
\end{enumerate}

From these assumptions, the main results of this study are presented in the following section.

\section{Optimization method for initialization variance}
\label{sec:reg_of_SGD}
 In ML, to obtain a high-performance model, we need to choose optimal hyperparameter settings such as the learning rate, batch size, and the properties of the initialization distribution. Building upon the continuous-time framework developed in Section \ref{sec:SDE_of_SGD}, this section investigates how the initialization variance $\sigma_0$ affects the final loss value in SGD. Our analysis begins by establishing the steady-state distribution of the weights via the Fokker–Planck equation. We also prove the relation between the variance of the steady state and the loss values by optimizing the variance of the initial distribution. 
 We first consider the KL divergence between the stationary distribution and the initial Gaussian density to relate the steady-state loss to the initial variance.
\begin{prop}\label{thm:KL_divergence}
Assume that the quasi-steady state distribution of Equation~\eqref{eq:ss} exists, and let $p_0$ be the Gaussian density of the initial distribution of the parameter in the SGD recurrence equation~\eqref{eq:Langevin}, i.e.,

\begin{equation}
    p_0(W)=\frac{1}{(2\sigma_0^2\pi)^{K/2}}e^{-\frac{\|W\|^2}{2\sigma_0^2}},
    \end{equation}
    where $K$ is the dimension of the parameter and $\sigma_0$ is the variance of the initial parameter $W^{(0)}$. Moreover, $L(W)$ is a cost function satisfying the quadratic form assumption with a deep local minimum. Then, the expected value of the loss, with respect to the steady-state distribution, satisfies the following inequality
        \begin{equation}
\begin{split}
\mathbb{E}_{\mathrm{ss}}\bigl[\bar{L}(W)\bigr]
&\le \frac{\alpha\,\sigma^2}{2\,b\,K}\Bigl[
       \tfrac{K}{2}\,\log\bigl(2\,\sigma_0^2\pi\bigr)
     + \log(C_{1,\mathrm{loc}})\\
&\quad
     + \frac{\mathbb{E}_{\mathrm{ss}}\bigl[\|W\|^2\bigr]}{2\,\sigma_0^2}
  \Bigr]\,.
\end{split}
 \label{eq:KL_div}
        \end{equation}
where $\mathbb{E}_{ss}$ refers to the expectation with respect to the steady-state distribution of the parameter $W$ and $\bar{L}(W)$ is the averaged loss function by the number of elements $K$.
        \begin{proof}
        We first recall the Fokker-Planck equation, the stationary state is given by
        \begin{equation}
        p_{\rm ss}(W)=C_{1,\text{loc}}e^{-\frac{2bL(W)}{\alpha\sigma^2}}. 
        \end{equation}
        Next, we introduce the Kullback–Leibler (KL) divergence between the initial state and the stationary state. The KL divergence is given by
        \begin{equation}
            D(p_{\rm ss}\|p_0)= \int p_{\rm ss}\log\left(\frac{p_{\rm ss}}{p_0}\right) dW\geq 0.
        \end{equation}
        Note that the KL divergence is always non-negative. Finally, substituting the initial distribution and the stationary state, we can obtain
        \begin{equation}
\begin{split}
\mathbb{E}_{\mathrm{ss}}\bigl[\bar{L}(W)\bigr]
&\le \frac{\alpha\,\sigma^2}{2\,b\,K}\Bigl[
       \tfrac{K}{2}\,\log\bigl(2\,\sigma_0^2\pi\bigr)
     + \log(C_{1,\text{loc}})\\
&\quad
     + \frac{\mathbb{E}_{\mathrm{ss}}\bigl[\|W\|^2\bigr]}{2\,\sigma_0^2}
  \Bigr]\,.
\end{split}
\end{equation}
        \end{proof}
\end{prop}
\begin{remark}
    By following Equation~\eqref{eq:KL_div}, we can consider the behavior of $\mathbb{E}_{\rm ss}[L(W)]$ within two different regions.
    \begin{itemize}
    \item When $\sigma_0^2\ll1$, the loss function behaves as 
    \begin{equation}
        \mathbb{E}_{\rm ss}[\bar{L}(W)]\leq K_1 \bar{V}(W),
        \label{eq:linear_relation}
    \end{equation}
    with $\mathbb{E}_{\rm ss}[W]=0$, $K_1$ is a constant and $\bar{V}(W)$ is the averaged variance of the steady-state weight distribution, which is divided by $K$. \item On a contrary, when $\sigma_0^2\gg1$, the expectation of the loss function behaves as
    \begin{equation}
        \mathbb{E}_{\rm ss}[\bar{L}(W)]\leq K_2 \log(\sigma_0^2).
    \end{equation}
    \end{itemize}
\end{remark}
In the small initial variance range, the behavior of the final state of the loss function depends on the final state of the weight variance $V(W)$. In this sense, when we set the small variances for the initialization, the value of the loss function is proportional to the size of the final state weight distribution. Meanwhile, when $\sigma_0\gg 1$, the value of loss function proportional to $\log(\sigma_0^2)$. Although at first glance the behavior under $\sigma_0^2 \gg 1$ might seem preferable, the large value of $\sigma_0^2$ causes the bound function to increase, offsetting the advantage. In particular, when the steady‐state variance $\mathbb{E}_{\rm ss}[\|W\|^2]$ remains on the same scale as $\sigma_0^2$, initializing with a smaller $\sigma_0^2$ yields a tighter loss bound than using a larger one.  Empirically, in the DNN model without vanishing‐gradient issues, the observed steady‐state variance indeed stays proportional to the chosen $\sigma_0^2$ (Section~\ref{sec:numerical}). Moreover, by optimizing the r.h.s of $\eqref{eq:KL_div}$ with respect to $\sigma_0$, we achieve the optimal condition for the initialization variance as stated in the following theorem.
\begin{theorem}\label{thm:optimized_KL}
    Under the assumptions of Proposition~\ref{thm:KL_divergence}, one has
        \begin{equation}
        \mathbb{E}_{\rm ss}[\bar{L}(W)]\leq \frac{\alpha \sigma^2}{4b}\left[\log\left( \frac{\mathbb{E}_{\rm ss}[\|W\|^2]}{K}\right) +\log(2\pi eC_{1,\mathrm{loc}})\right],
    \end{equation}
    provided that the optimal standard deviation of the initial distribution is given by
\[
\sigma_0 = \sqrt{\frac{\mathbb{E}_{\rm ss}[\|W\|^2]}{K}}=\sqrt{\bar{V}(W)}.\label{eq:optimality}
\]
If we assume $\mathbb{E}_{ss}[W]=0$, the optimized bound for the expected value of the loss function is:
        \begin{equation}
        \mathbb{E}_{\rm ss}[L(W)]\leq \frac{\alpha \sigma^2}{4b}\left[\log(\bar{V}(W) ) +\log(2\pi eC_{1,\mathrm{loc}})\right],
    \end{equation}
where $\bar{V}(W)$ is the averaged variance of the final state. 
\end{theorem}
The direct consequence of our result provides that $\sigma_0=\sqrt{\mathbb{E}_{\rm ss} [|W|^2 ]/K}$, where $K$ is the total number of units of the network. For a given structure of the network, the optimal scale of the initialization standard deviation is $O(1/\sqrt{K})$. This result is also consistent with He-normal distribution where the initialization standard deviation at layer $\ell$ is of order $O(1/\sqrt{K_\ell})$, where $K_\ell$ is the size of the layer. 
\begin{remark}
  The inequality in Theorem~\ref{thm:optimized_KL} depends on the normalizing constant $C_{1,\mathrm{loc}}$, which captures the information about the loss landscape and model structure. A common way to approximate $C_{1,\mathrm{loc}}$ is to apply the Laplace approximation for each local minimum and sum the resulting Gaussian normalizers, as mentioned in, for instance, \cite{gelman1995bayesian,jastrzkebski2017three}. In our theorem, $C_{1,\mathrm{loc}}$ should be relatively smaller than $\bar{V}$. If we consider a one-layer model, the normalized constant is larger than $C_{1,\mathrm{loc}}$ because it has only a single minimum.
\end{remark} 

\begin{remark}
  According to \cite{mandt2017stochastic}, under the local quadratic approximation \eqref{eq:quasi_loss}, the quasi-steady-state distribution is approximated by a Gaussian
  \begin{equation}
      \tilde{p}_{\rm ss}\propto e^{-\frac{1}{2}W^T C^{-1}W},
  \end{equation}
  where $C$ is the quasi stationary covariance of $\tilde{p}_{\rm ss}$. Note that for a general constant anisotropic noise covariance $\Sigma$, the quasi-steady state is no longer a Gibbs-type density. With the normalization, we set the quasi-steady state to be
  \begin{equation}
      \tilde{p}_{\rm ss}(W)=\frac{1}{(2\pi)^{K/2} \det(C)^{1/2}}e^{-\frac{1}{2}W^T C^{-1}W}.
  \end{equation}
  In addition, this covariance matrix satisfies the following relation
  \begin{equation}
      AC+CA^T=\ep \Sigma,
  \end{equation}
  where $A$ is already introduced in \eqref{eq:quasi_loss}. Then, the KL divergence is calculated by
\begin{equation}
\begin{aligned}
&D_{\rm KL}(\tilde{p}_{\rm ss}\|p_0)\\&
=\frac{1}{2}\Bigg[
-\log\det(C)-K+K\log\sigma_0^2
+\frac{\mathbb{E}_{\rm ss}\!\left[\|W\|^2\right]}{\sigma_0^2}
\Bigg].
\end{aligned}
\end{equation}
  Moreover, we can optimize the KL divergence to obtain an optimal condition for initialization variance $\sigma_0^2$. It is given by
  \begin{equation}
      \sigma_0=\sqrt{\frac{\mathbb{E_{\rm ss}}[\|W\|^2]}{K}}.
  \end{equation}
  It is the same expression as Theorem. \eqref{eq:optimality} but $\mathbb{E}[|W|^2]=\tr (C)$ with $\mathbb{E}_{\rm ss}[W]=0$. If $C$ is an isotropic covariance, this result corresponds to the theorem. \ref{thm:optimized_KL}. For the anisotropic constant covariance case, we take into account $\tr (C)$. In this setting, we do not obtain an explicit bound for $\mathbb{E}_{\rm ss}[L(W)]$ as in the isotropic case. However, $L(W)$ and the KL divergence are related through the quasi-steady covariance $C$, and by optimizing the KL divergence, it still provides an optimal initialization condition $\sigma_0^2=\tr(C)/K$.
\end{remark} 

In summary, our analysis shows that the expected loss is highly sensitive to the scale of the initial weight distribution. In particular, for small initialization variances, the loss is (at most) proportional to $\bar{V}(W)$. In this sense, to obtain the small loss values, we need to suppress the variances of the steady state. Ideally, if one can set the initialization variance to $\sigma_0=\sqrt{\bar{V}(W)}$, which implies that the initial distribution is very close to the steady state, the optimized loss bound is achieved. However, satisfying this optimal condition before the learning process is very challenging in practice. In the following numerical section, we further explore the relationship between $\mathbb{E}_{\rm ss}[\bar{L}(W)]$ and $\bar{V}(W)$ and discuss the practical difficulties in meeting this optimal condition.

\section{Numerical experiments}
\label{sec:numerical}
\subsection{Setup}
In this section, we perform the simulation on the Deep Neural Network (DNN) model. Our model is a basic DNN model, a fully connected neural network comprising an input layer, hidden layers, and an output layer. The NN model takes training data of size $N$, $(x_i,y_i)$ for $i = 1,2,\ldots,N$. Throughout this article, the input $x_i=(x_{i1},\ldots,x_{id})$ is normalized to be in $[0,1]^d$, and the output $y_i\in\mathcal{Y}=\{1,\ldots,M\}$, where $M$ is the number of classes of the classification task. In this case, the weight $W=(W_{jm})$ is a $d\times M$ matrix, connecting the input layer to the output layer.
To perform classification, we use the softmax function to normalize output values. Moreover, for any input $x_i$, the transformed value is defined by $z_i=x_iW=(\sum_{j=1}^dx_{ij}W_{j1},\ldots,\sum_{j=1}^dx_{ij}W_{jM})\in\mathbb{R}^M$. Then, the soft-max function evaluated at $z_i$ is $\phi(z_i)=(\phi_1(z_i),\ldots,\phi_M(z_i))$ and is defined by
\begin{equation}
	\phi_{m}(z_i)=\frac{e^{z_{im}}}{\sum_k e^{z_{ik}}},\text{ for }m=1,\ldots,M.
	\label{eq:softmax}
\end{equation}

In this study, the loss function is chosen as the cross-entropy. In other words, the loss function of the weight $W$ evaluated at any input $x$ is defined by,

\begin{equation}
	L_{\rm cross}(W;x)= -\sum_{m=1}^M t_{m}\log \phi_m(xW),
	\label{eq:cross}
\end{equation}
where $t=(t_1,\ldots,t_M)$ is the one-hot coding of the truth label $y$, i.e.,

\begin{equation}
\label{eq:one_hot_coding}
    y=m_0 \Leftrightarrow t_m= \begin{cases}
         0,&\mbox{if }m \neq m_0\\
        1&\mbox{if }m = m_0
    \end{cases}.
\end{equation}
In addition, for minibatch learning, the global loss function is given by,
\begin{equation} 
	\tilde{L}(W)=\frac{1}{b}\sum_{i\in\Omega_b}L_{\rm cross}(W;x_i),
\end{equation}
where $\Omega_b \subset \{1,\ldots,n\}$ is a random subset of size $b$. Then, the SGD optimization step is written as 
\begin{align}
	W^{(t+1)}_{jm}=W^{(t)}_{jm}-\frac{\alpha}{b}\sum_{i\in\Omega_b}\frac{\p {L_{\rm cross}}(W^{(t)};x_i)}{\p W_{jm}}, \non\\ \text{ for }j=1,\ldots,d, \text{ and }m = 1,\ldots,M,
	\label{eq:sgd}
\end{align}
for minibatch learning. In our problem, the optimization method on an NN model with no middle layer is based on the above setup and iterative updates following the SGD optimization algorithm.
\subsection{Numerical results}
We carry out classification experiments on MNIST with a three-layer fully connected network with ReLU activations.  All weights are initialized as $W_0\sim\mathcal N(0,\sigma_0^{2})$ and the models are trained with stochastic gradient descent (SGD).  We aim to examine the relationship between the expected value of the steady-state loss and the variance of the steady-state weight distribution.

In the three-layer network, the loss function is non-convex, multiple minima coexist, and the theorem predicts a non-trivial dependence between the final loss and the variance of the initial distribution $\sigma_0^2$. In what follows, we present numerical results that primarily validate Theorem~\ref{thm:KL_divergence} and, in addition, illustrate the time evolution of the weight variance. In our theoretical result, we set \(V(W):=\mathbb{E}[\|W\|^{2}]\) under the symmetry assumption \(\mathbb{E}[W]=0\). In our simulations, the empirical mean satisfies \(\|\mathbb{E}[W]\|^{2}\ll\mathbb{E}[\|W\|^{2}]\); hence the numerical variance can be compared directly with the theoretical quantity \(\bar{V}(W)\). In this numerical experiment, some hyperparameters are fixed, including batch size $b=100$, the learning rate $\alpha=0.0001$, and the number of epochs $t=1000$ for all simulations. Each case is performed 10 times independently, and the averaged loss function $\mathbb{E}_{ss}[\bar{L}(W)]$ is computed from these results. The experiments are conducted using two different settings: (1) A simple three-layer DNN model trained on the MNIST dataset, and (2) CNN-based models including a two-layer CNN and a VGG model \cite{simonyan2014very} trained on the CIFAR-10 dataset.

\subsubsection {Three-layer DNN model}
\begin{figure}[htbp]
  \centering
\subfloat[Time evolution of the averaged weight variance.]{
  \label{fig:time_evolution_var_DNN_a}
  \includegraphics[width=0.45\textwidth]{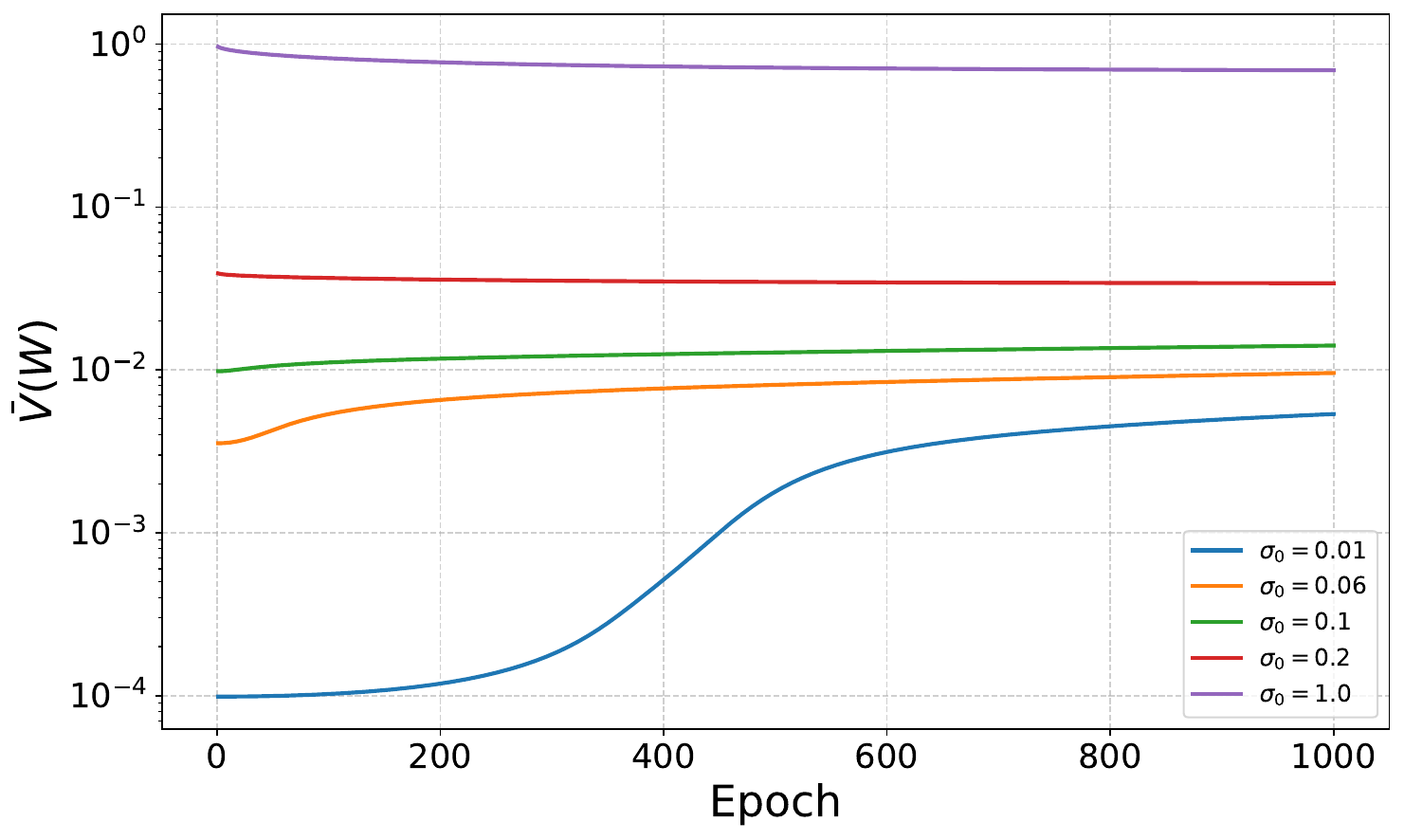}
}
  \hfill
\subfloat[Validation accuracy for different $\sigma_0$ values.]{
  \label{fig:accuracy-comparison_DNN}
  \includegraphics[width=0.45\textwidth]{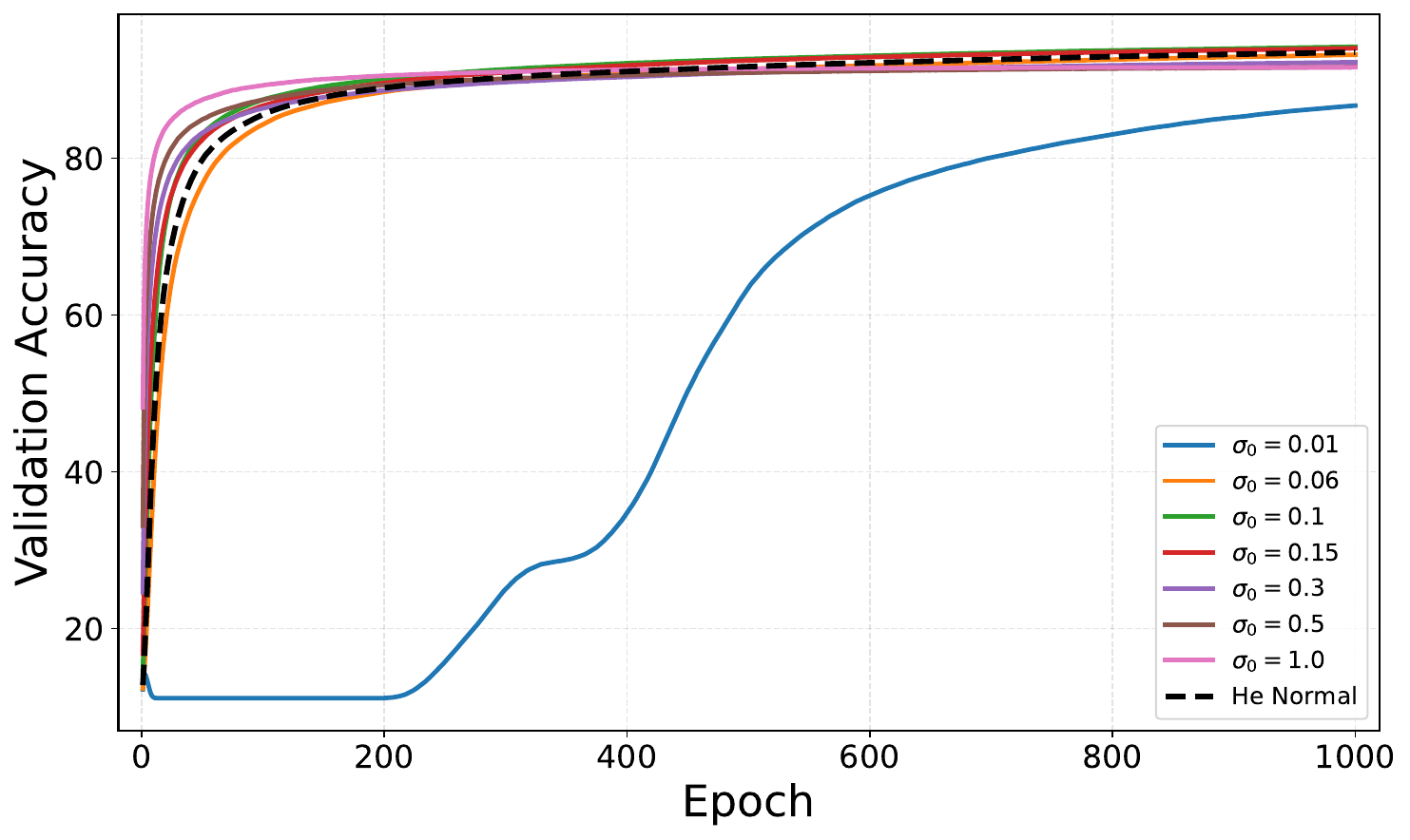}
}
  \caption{(a) Time evolution of the averaged weight variance \(V(W,t)\) over $N=10$ runs of a three-layer deep neural network trained with SGD under different initial variances \(\sigma_0\). Results are shown for MNIST. (b) The figure shows the accuracy for various \(\sigma_0\) values on MNIST. When \(\sigma_0=0.15\), the validation accuracy takes the highest value. The black dashed line represents the result using He-Normal initialization.}
  \label{fig:time_evolution_var_DNN}
\end{figure}

In this section, we show the numerical results for a deep neural network model with three layers. First, we show the results of the time evolution of the averaged variance. In Figure~\ref{fig:time_evolution_var_DNN}\subref{fig:time_evolution_var_DNN_a}, by examining the time evolution of these variances in the three‐layer DNN, we observe in this case that when \(\sigma_0 \le 0.1\), the steady‐state variance at the end of the training epoch is larger than the initial variance, whereas for \(\sigma_0 \ge 0.2\), the final variance falls below the initial value. Intuitively, the DNN appears to adjust its weights throughout the training process to converge toward a locally optimal variance.

\begin{figure}[htbp]
  \centering
\subfloat[Overall variance for the network.]{
  \label{fig:loss_vs_variance_DNN_a}
  \includegraphics[width=0.45\textwidth]{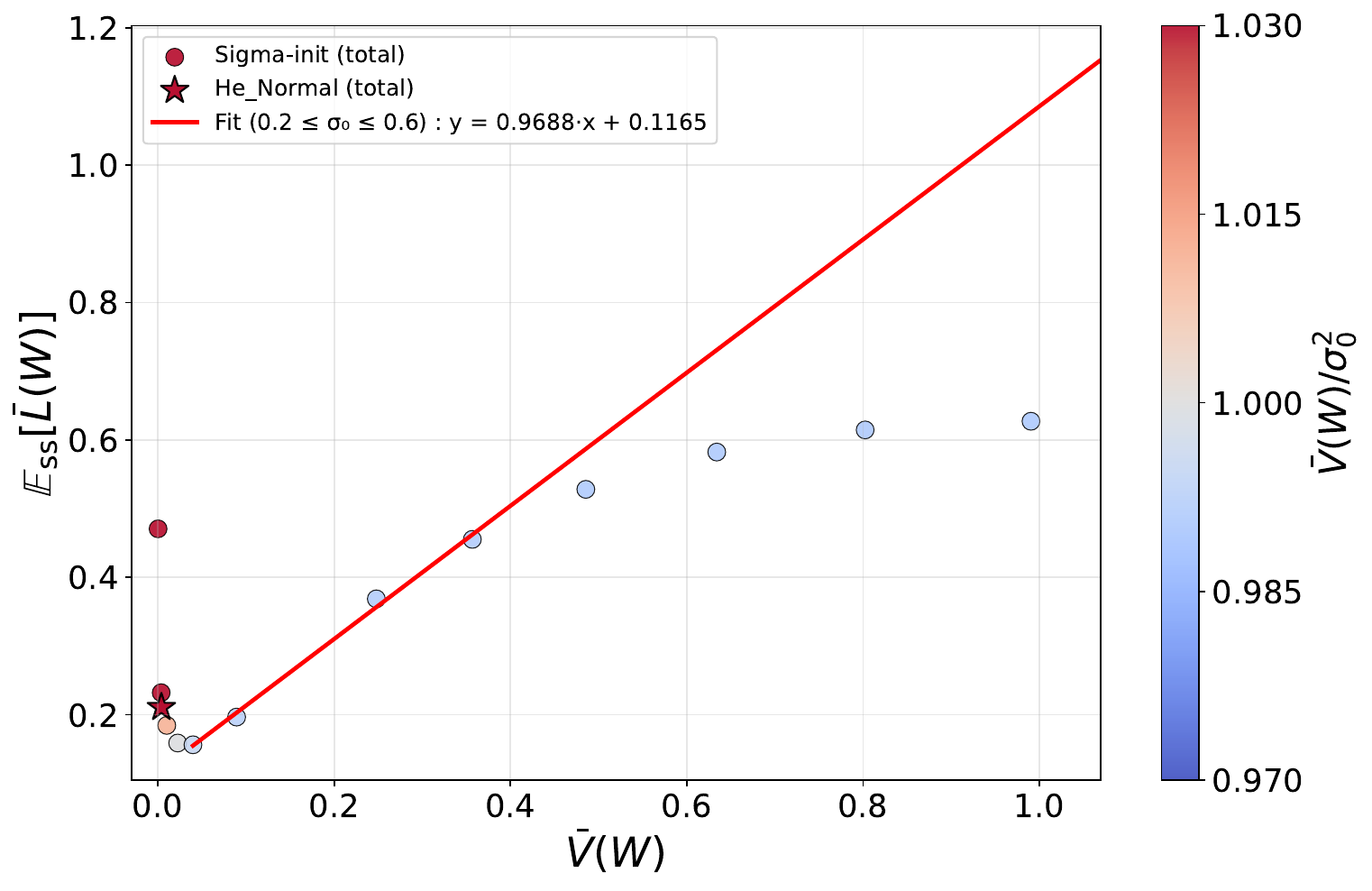}
}

\subfloat[Variance of the output layer.]{
  \label{fig:loss_vs_variance_DNN_b}
  \includegraphics[width=0.45\textwidth]{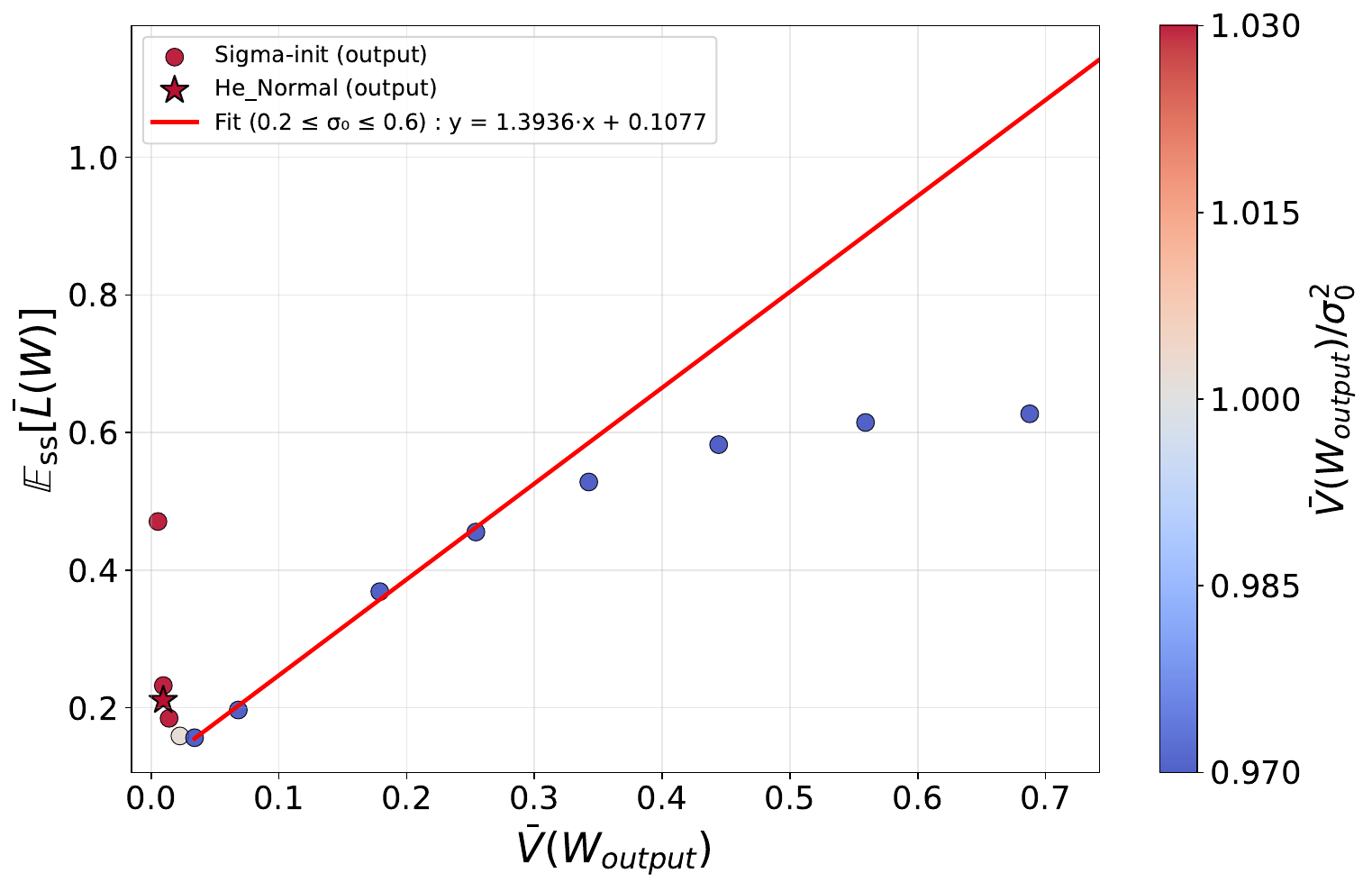}
}
    \caption{
    The relationship between the expected training loss \(\mathbb{E}[L(W)]\) and the steady‑state variance \(\bar{V}(W)\) in Equation~\eqref{eq:linear_relation} for a DNN network trained using SGD under different initial variances \(\sigma_0^2\). The red linear function is the most relaxed bound for small $\sigma_0$ shown in Equation \eqref{eq:linear_relation}, and a star marker represents the result using He-Normal initialization. For each \(\sigma_0\), weights are independently sampled from the same Gaussian distribution \(\mathcal{N}(0,\sigma_0^2)\), and the average variance $\bar{V}(W)$ over 10 runs is plotted on the x‑axis with the MNIST dataset. The figures show (a) the overall variance for this network, (b) the variance of the output layer. Each point is colored according to the ratio $\bar{V}(W)/\sigma_0^2$. }
  \label{fig:loss_vs_variance_DNN}
\end{figure}

 Note that for tiny values of $\sigma_0$, the steady-state variance $\bar{V}(W)$ seems to violate the bound of Equation~\eqref{eq:linear_relation}.
 This issue likely stems from the \textit{vanishing gradient problem}, a common challenge in training deep neural networks using too small initial weights. This is a technical problem; thus, we have to discuss our theoretical analysis without this region. In Figure \ref{fig:loss_vs_variance_DNN}\subref{fig:loss_vs_variance_DNN_a}, with a reasonably small region of $\sigma_0$ implemented on MNIST data, the result shows that the optimization process gives a lower loss value than a linear bound $K_1\bar{V}(W)$ on the right-hand side of \eqref{eq:linear_relation}. In this sense, \eqref{eq:linear_relation} holds for this numerical simulation. In particular, the optimized $\sigma_0$, the one satisfying the optimal condition \eqref{eq:optimality}, provided a slightly lower loss function than the case of using He-normal initialization, which is a common initialization distribution.

Figure~\ref{fig:loss_vs_variance_DNN}\subref{fig:loss_vs_variance_DNN_a} highlights the ratio \(\bar{V}(W)/\sigma_0^2\) as a function of the overall converged variance $\bar{V}(W)$. At \(\sigma_0 = 0.15\), this ratio approaches unity, which corresponds to the point in Figure~\ref{fig:loss_vs_variance_DNN}\subref{fig:loss_vs_variance_DNN_a} (loss vs variance of overall layer) and \ref{fig:loss_vs_variance_DNN}\subref{fig:loss_vs_variance_DNN_b} (loss vs variance of the output layer) attaining the minimum expected loss. Hence, satisfying \(\sigma_0^2 = \bar{V}(W)\) realizes the lowest loss, which again confirms our theoretical result.

Finally, we illustrate the time evolution of the validation accuracy for a learning process with our three-layer DNN model. Figure~\ref{fig:time_evolution_var_DNN}\subref{fig:accuracy-comparison_DNN} shows that when the optimal $\sigma_0=0.15$ is selected for the initialization, the model achieves the highest accuracy in the final epoch. Furthermore, the trajectories corresponding to the standard deviations around this optimal $\sigma_0$ also achieve better accuracies than the case of He normal initialization.
In summary, our numerical experiments for the DNN network under SGD confirm the following key points:
\begin{itemize}
    \item In Figure~\ref{fig:time_evolution_var_DNN}\subref{fig:time_evolution_var_DNN_a}, we observe that the learning dynamics tend to converge toward the ideal scale of the weight distribution. Specifically, the algorithm dynamically modulates the spread of the weight distribution by adapting the scale of the parameters around the optimal $\sigma_0$ to align itself with the theoretically optimal size.
    \item For initial variances in the range \(\sigma_0^2 \le 1\), the expected value of the loss function satisfies our bound of Equation~\eqref{eq:linear_relation}:
   \[
     \mathbb{E}_{\rm ss}[L(W)] \leq K_1\bar{V}(W)
   \]
This bound is the most relaxed bound function for the small $\sigma_0$ region.
\item The theory predicts that the optimal initialization should satisfy
   \[
     \sigma_0^2 = \bar{V}(W).
   \]
   In practice, DNN models are capable of achieving this optimal condition if they start from an ideal scale of the initialization distribution, especially in the Gaussian case. In particular, although the He-normal scheme remains a strong baseline, our results show that an alternative initialization variance $\sigma_0$, derived from the proposed optimal condition, achieves even better training performance, particularly in terms of lower loss, than He-normal initialization. Physically, this optimal condition means that SGD begins its search in exactly the region of parameter space that will eventually contain the minimum-loss solution.  Because the exploration range is already well-matched to the final steady-state spread, every weight is updated inside this region from the outset, enabling the network to converge more quickly and reliably to well-optimized values.
   \item While the theoretical optimal condition is expressed in terms of the overall ratio \(\bar{V}(W)/\sigma_0^2 = 1\), our numerical experiments suggest that lower loss is achieved when this condition is also approximately satisfied in individual layers. In particular, Figure~\ref{fig:loss_vs_variance_DNN}\subref{fig:loss_vs_variance_DNN_b} indicates that the variance ratio of the output layer affects the loss. This suggests that satisfying the optimal condition is important not only for the overall variance but also for each individual layer.
\end{itemize}
\subsubsection{Modern network models}
In this section, we present numerical results for more realistic architectures trained on the CIFAR-10 dataset. 
\begin{figure}[htbp]
  \centering
\subfloat[Two-layer CNN model.]{
  \label{fig:loss_vs_variance_CIFAR10_a}
  \includegraphics[width=0.45\textwidth]{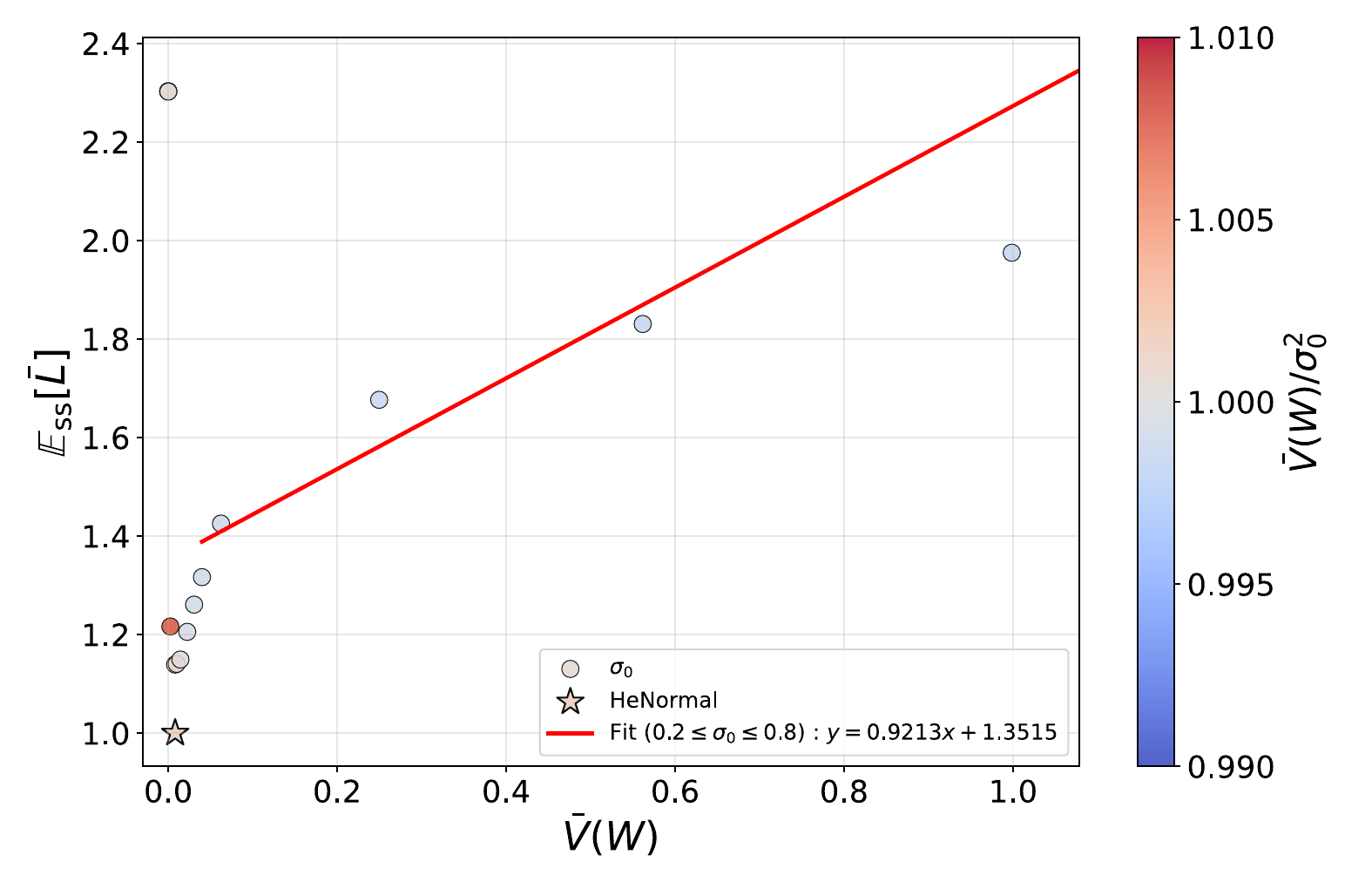}
}

  \hfill
\subfloat[VGG model.]{
  \label{fig:loss_vs_variance_CIFAR10_b}
  \includegraphics[width=0.45\textwidth]{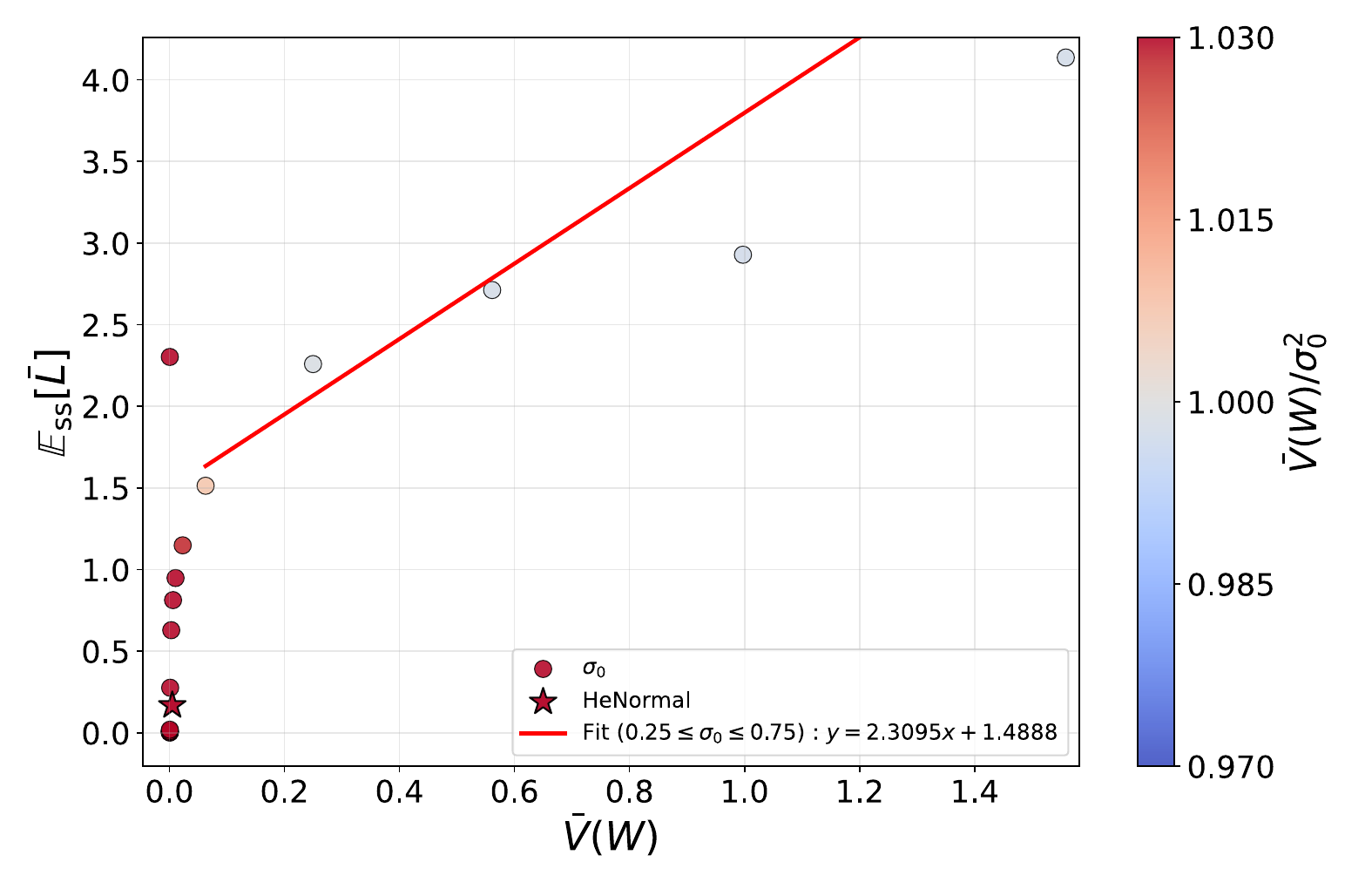}
}
    \caption{The relationship between the expected training loss \(\mathbb{E}[L(W)]\) and the steady‑state variance \(\bar{V}(W)\) in Equation~\eqref{eq:linear_relation} for a modern complex network trained using SGD under different initial variances \(\sigma_0^2\). The red linear function is the most relaxed bound for small $\sigma_0$ shown in Equation \eqref{eq:linear_relation}, and a star marker represents the result using He-Normal initialization. For each \(\sigma_0\), weights are independently sampled from the same Gaussian distribution \(\mathcal{N}(0,\sigma_0^2)\), and the average variance $\bar{V}(W)$ over 10 runs is plotted on the x‑axis with the CIFAR-10 dataset. The figures show (a) the CNN two-layer model and (b) the VGG Net model. Each point is colored according to the ratio $\bar{V}(W)/\sigma_0^2$.
  }
    \label{fig:loss_vs_variance_CIFAR10}
\end{figure}

Figure~\ref{fig:loss_vs_variance_CIFAR10}\subref{fig:loss_vs_variance_CIFAR10_a} shows the results for the two-layer CNN. In this case, He-normal initialization gives the lowest loss among the tested initializations. This appears to be closely related to the fact that the variance ratio \(\bar{V}_\ell(W)/\sigma_{0,\ell}^2\) in He-normal initialization remains close to unity across all layers, including the convolutional layers. In this sense, He-normal provides an initialization scale that is already well aligned with the practically favorable condition observed in our earlier experiments.
By contrast, the VGG results in Figure~\ref{fig:loss_vs_variance_CIFAR10}\subref{fig:loss_vs_variance_CIFAR10_b} do not exhibit the same trend as in the DNN and two-layer CNN cases. In particular, the loss is no longer organized simply by the ratio \(\bar{V}(W)/\sigma_0^2\), and the relationship predicted by the theory is not clearly observed for this model. This may indicate that, for a deeper and more complex architecture such as VGG, the assumptions underlying our analysis are no longer sufficiently reasonable. In particular, our theoretical argument relies on a local quadratic approximation of the loss function around a local minimum, whereas such an approximation may not well describe the optimization landscape relevant to VGG training.

\section{Conclusion}
\label{sec:conclusion}
In this article, we derived an analytic optimal condition for the initial variance \(\sigma_0^2\) of Gaussian weight initialization under SGD and examined it numerically using a fully connected three-layer neural network model on MNIST and modern convolutional architectures on the CIFAR-10 dataset. For the three-layer model, our experiments showed that the theoretically optimal \(\sigma_0\) consistently achieves lower training loss and higher test accuracy than the widely used He-normal initialization. For the two-layer CNN model, He-normal initialization gave the best performance among the initializations tested in our experiments, which appears to be related to the fact that it satisfies the optimal variance condition well in this case. In particular, our results indicate that He-normal satisfies the optimal condition at each layer, and this provides a mathematical perspective on why He-normal initialization performs well in practice. In some shallow DNN models, the loss function structure usually consists of multiple quadratic local minima, and therefore, the learning process is practically non-ergodic. In this sense, studying the ideal condition for the initialization is important because the influence of the initial state persists in the steady state, and starting the learning process from an ideal situation likely yields the most efficient result. By contrast, for the VGG model, we observed that the predicted variance-loss relation is no longer clearly observed. This suggests that, in deeper and more complex architectures, the assumptions underlying our analysis, in particular the local quadratic approximation around a local minimum, may no longer provide a sufficient description of the effective optimization landscape. Physically, choosing the initialization distribution is equivalent to choosing the exploration region in the learning process, especially for small learning rates. This optimal condition means that the exploration region of SGD already matches the region where the final state is located. Every weight is updated inside a productive basin from the first step, yielding more reliable optimization in landscapes riddled with multiple local minima. Although a lower training loss does not always translate into the highest test accuracy, our experiments show that the optimal Gaussian parameter $\sigma_0$ improves both metrics in practice. It is still meaningful for the process to achieve the lowest loss value in ML framework.

While previous works showed a few mathematically grounded criteria for choosing initialization hyperparameters, our method provides a concrete guideline by evaluating the information distance between initialization and steady state in a learning process. For future perspective, discovering algorithms that actively search for the optimal initialization variance could further improve the efficiency of deep learning optimization. Moreover, although our current analysis relies on the Kullback–Leibler divergence, replacing it with metrics such as the Wasserstein distance would naturally recast the problem within the framework of optimal transport and may lead to richer theoretical insights.
Moreover, we can also consider the heavy-tailed distribution for initialization. This aspect has been studied, for example, by \cite{gurbuzbalaban2021fractional}. However, in our theorem, heavy-tailed initializations do not deliver uniformly high accuracy across trials. Indeed, their divergent second moment often makes performance less stable than in the Gaussian case. Nevertheless, they sometimes yield surprisingly good results in individual runs, suggesting that, under the right conditions, the broader exploration they induce can land the optimizer in favorable regions in the loss function space. Identifying when and why these rare successes occur remains an interesting area for future work.

\newpage
\bibliographystyle{IEEEtran}
\bibliography{NN}


\end{document}